\def\year{2021}\relax
\documentclass[letterpaper]{article} 
\usepackage{aaai21}  
\usepackage{times}  
\usepackage{helvet} 
\usepackage{courier}  
\usepackage[hyphens]{url}  
\usepackage{graphicx} 
\usepackage{natbib}  
\usepackage{caption} 

\usepackage{multicol}
\usepackage{multirow}
\usepackage{verbatim}
\usepackage{amsmath} 

\title{A Hierarchical Reasoning Graph Neural Network\\ for The Automatic Scoring of Answer Transcriptions in Video Job Interviews}
\author{
    Kai Chen, Meng Niu, Qingcai Chen$^\star$

}
\affiliations{
    Harbin Institute of Technology (Shenzhen)

}

\begin{document}


\maketitle

\begin{abstract}

We address the task of automatically scoring the competency of candidates based on textual features, from the automatic speech recognition (ASR) transcriptions in the asynchronous video job interview (AVI). The key challenge is how to construct the dependency relation between questions and answers, and conduct the semantic level interaction for each question-answer (QA) pair. However, most of the recent studies in AVI focus on how to represent questions and answers better, but ignore the dependency information and interaction between them, which is critical for QA evaluation. In this work, we propose a Hierarchical Reasoning Graph Neural Network (HRGNN) for the automatic assessment of question-answer pairs. Specifically, we construct a sentence-level relational graph neural network to capture the dependency information of sentences in or between the question and the answer. Based on these graphs, we employ a semantic-level reasoning graph attention network to model the interaction states of the current QA session. Finally, we propose a gated recurrent unit encoder to represent the temporal question-answer pairs for the final prediction. Empirical results conducted on CHNAT (a real-world dataset) validate that our proposed model significantly outperforms text-matching based benchmark models. Ablation studies and experimental results with 10 random seeds also show the effectiveness and stability of our models.

\end{abstract}

\section{Introduction}

\noindent Recent years have witnessed the rapid advancement of online recruitment platforms. With the increasing amount of online recruitment data, more and more interview related studies have emerged such as person-job (or talent-job) fit~\cite{shen2018joint, qin2018enhancing,luo2019resumegan, bian2019domain} and automatic analysis of asynchronous video interviews (AVIs)~\cite{hemamou2019hirenet, hemamou2019slices,suen2019tensorflow}, which aim to enable automated job recommendation and candidate assessment. Among these studies, person-job fit is to casting the task as a supervised text match problem. Given a set of labeled data (i.e., person-job match records), it aims to predict the matching label between the candidate resumes and job description. More recently, deep learning has enhanced person-job fit methods by training more effective text match or text representations models~\cite{xu2017measuring, jiang2019semantic}. AVI is to determine whether the candidate is hirable by evaluating the answers of interview questions. In AVIs, an interview is usually considered as a sequence of questions and answers containing salient socials signals. To evaluate the candidates more comprehensively, AVI models will extract the features of video (or image), text, and voice in the process of answering questions. In this work, we focus on the scoring of multiple QA pairs,  we only extract the features of text modality and define this task as the scoring competency of candidates rather than the score of whether or not to be employed.
Based on the anatomy of the human interviewers' evaluation process, the solutions consist of two stages: (1) analyzing and evaluating individual QA pair one by one, then acquiring the evaluation status, and (2) grading the competency of the candidate based on the evaluation status of multiple QA pairs. 

For the first stage, existing methods tend to employ text matching or attentional text matching algorithms to evaluate QA pairs~\cite{hemamou2019hirenet, suen2019tensorflow}, which feeds the concatenated representation of the question and the answer to the subsequent classifier. As we all know, questions in an asynchronous video interview are not limited to specific domains. That is to say, candidates can answer questions according to their work or study experience. In this way, the candidates' answers will be varied and it is difficult to evaluate the answer accurately only by text matching. Intuitively, it is more reasonable to evaluate QA pairs through the semantic interaction between questions and answers. A critical challenge along this line is how to reveal the latent relationships between each question and answer.

Graph neural networks (GNNs, ~\cite{dai2016discriminative, yao2019graph, ghosal2019dialoguegcn}) can learn effective representation of nodes by encoding local graph structures and node attributes. Due to the compactness of model and the capability of inductive learning, GNNs are widely used in modeling relational data~\cite{battaglia2018relational, schlichtkrull2018modeling, pan2020semantic} and logical reasoning~\cite{luo2019improving, jiang2020reasoning}. Recently, ~\citet{zhang2020efficient} proposed a GNN variant, Named ExpressGNN, to strike a nice balance between the representation power and the simplicity of the model in probabilistic logic reasoning.~\citet{ghosal2019dialoguegcn} constructed the DialogeGCN to address context propagation issues present in the RNN-based methods. Specifically, they leverage self and inter-speaker dependency of the interlocutors to model conversational context for emotion recognition. Inspired by~\cite{ghosal2019dialoguegcn}, we present a sentence-level relational GCN to represent the internal temporal and QA interaction dependency in the process of answering questions. 




For the second stage of grading the candidate, based on the representation of QA pairs, exists methods~\cite{hemamou2019hirenet,hemamou2019slices} prefer to encoder question-answer pairs as a sequence directly. However, this kind of approaches lead to insufficient interaction between the semantic information of question and answer pairs. Therefore, it is difficult to ensure the rationality and explainability of the evaluation. To mitigate this issue, in the first stage, we present a semantic-level graph attention network (GAT) to model the interaction states of each QA session.


To this end, we propose a Hierarchical Reasoning Graph Neural Network (HRGNN) for the automatic scoring of answer transcriptions (ASAT) in job interviews. Specifically, the proposed sentence-level relational graph convolutional neural network (RGCN) is used to capture the contextual dependency, and the semantic-level Reasoning graph attention network (RGAT) is applied to acquire the latent interaction states. And the contribution of our work can be summarized as follows:
\begin{itemize}
\item We propose a relational graph neural network to remedy the lack of QA interaction in previous assessment methods. Specifically, the relation of internal temporal dependency in each question/answer is helpful for context understanding. And the relation of QA interaction dependency can establish the latent semantic interaction of question-to-answer and answer-to-question.
\item To our knowledge, we are the first one to construct a hierarchical graph neural network for ASAT to model the relation between sentences in the question and its homologous answer, and interact and infer at the semantic level. Although we just evaluate the competence of candidates on textual features, it greatly improves the rationality and accuracy of the evaluation.

\item Our model can outperform all existing benchmark approaches on a Chinese real-world dataset. Ablation studies and experimental results with 10 random seeds show the effectiveness and stability of our models.
\end{itemize}

\section{Related Work}
{\bf Asynchronous Video Interviews} The asynchronous video interview is considered as one of the most essential tasks in talent recruitment, which forms a bridge between employers and candidates in fitting the eligible person for the right job~\cite{shen2018joint,hemamou2019slices,hemamou2019hirenet}. ~\citet{shen2018joint} developed a joint learning system to model job description, candidate resume, and interview assessment. It can effectively learn the representation perspectives of the different job interview process from the successful job interview records and then applied in person-job fit and interview question recommendation. ~\citet{hemamou2019hirenet} takes an interview process as a sequence of questions and answers and proposed a hierarchical attention model named HireNet to predict the hireability of the candidates. As far as we know, these approaches ignore the deep dependency between interview questions and answers.

\noindent {\bf Short Answer Scoring} Automatic short answer scoring (ASAS)~\cite{claudia2003automated,sultan2016fast,lun2020multiple, goenka2020esas} is a research subject of intelligent education, which is a hot field of natural language understanding. Methods for ASAS are driven with the help of deep learning techniques~\cite{mueller2016siamese,zhao2017memory} and domain-specific knowledge~\cite{conneau2017supervised, goenka2020esas}. Recently,~\citet{saha2019joint} have used InferSent~\cite{conneau2017supervised} and neural domain adaptation to obtain state-of-the-art results in the ASAS task. ~\citet{lun2020multiple} proposed multiple data augmentation strategies to learn language representation and achieved a significant gain over benchmark models. It should be emphasized that, in ASAS tasks, the answer text is short and the domain is specific. For the ASAT task which contains several open-domain interview questions, the scoring of long-text answers is much more challenging. 

\noindent {\bf Graph Neural Network} Graph neural networks have been successfully applied to several natural language processing tasks, such as text classification~\cite{velivckovic2017graph, yao2019graph, zhang2018sentence, zhang2020every}, machine translation~\cite{marcheggiani2018exploiting}, question generation~\cite{pan2020semantic,chen2020toward} and fact verification~\cite{zhou2019gear}. ~\citet{zhou2019gear} propose a graph-based evidence aggregating and reasoning (GEAR) framework which enables information to transfer on a fully-connected evidence graph and then utilize different aggregators to collect multi evidence information.~\citet{pan2020semantic} constructed a semantic-level graph for content selection and improved the performance over questions requiring reasoning over multiple facts. Inspired by previous researches, we proposed a hierarchical reasoning graph neural network to alleviate the issues of lacking interaction and semantic reasoning between questions and answers in the video job interview.

\section{Methods}
\begin{figure*}[!ht]
\centering
\centerline{ {\includegraphics[width=1.0\textwidth]{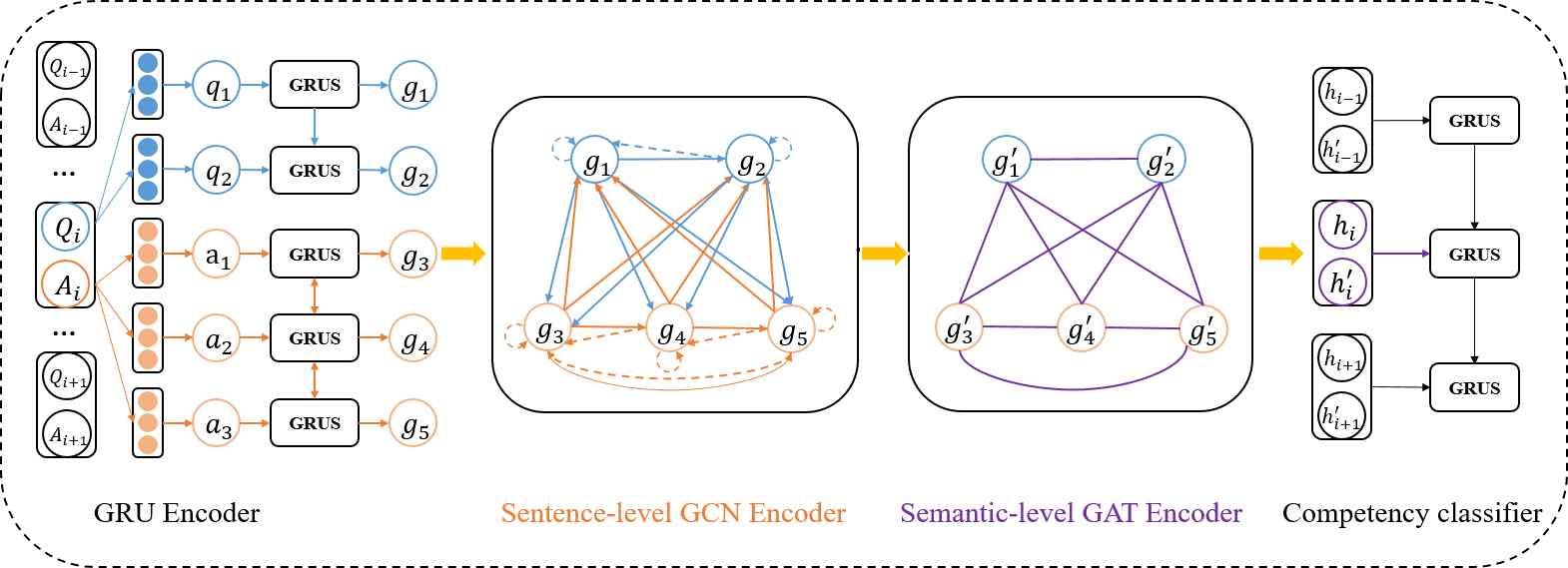}}}
\caption{An overview of the proposed HRGNN. As an example, upon a QA session in the interview, every sentence node updates itself from its neighbors. In the sentence-level GCN, each updated node contains dependency information and will be aggregate by the subsequent GAT encoder on the semantic-level. Finally, a GRU network is employed to represent the global interview for competency prediction.}
\label{fig-arc}
\end{figure*}

We now introduce the proposed Hierarchical Reasoning Graph Neural Network (HRGNN) for the automatic scoring of answer transcriptions in the job interview. HRGNN consists of four integral components --- Gated Recurrent Unit Encoder, Sentence-level GCN Encoder, Semantic-level GAT Encoder, and Competency Classifier. An overview of HRGNN is shown in Fig.~\ref{fig-arc}.
We first give some detailed explanation about Problem Formalization.

\subsection{Problem Formalization}
We represent the automatic scoring of answer transcriptions in video job interview as an object composed of ${n}$ question-answer pairs ${\{\{Q_1,A_1\},\{Q_2,A_2\},...,\{Q_n,A_n\}\}}$. In our model, the ${i}$\-th question ${Q_i}$ is a sequence of ${c_{Q_i}}$ sentences (sub-questions) ${\{q_1^i,q_2^i,...,q^i_{c_{Q_i}}\}}$, where ${c_{Q_i}}$ represents the count of sentences in question ${i}$. Subsequently, the ${j}$-th sentence ${q_j}$ in the question $Q_i$ can be formulated as $\{{w^{q_j}_1,w^{q_j}_2,...,w^{q_j}_{l_{q_j}}}\}$. Here ${l_{q_j}}$ is the number of words in sentence $j$. ${A_i}$ denotes the sequence of words ${\{a_1^i,a_2^i,...,a^i_{c_{A_i}}\}}$ describing the ${i}$\-th answer. And ${c_{A_i}}$ denotes the number of sentences in the ${i}$\-th answer. In a same way, the ${j}$-th sentence ${a_j}$ in the answer $A_i$ is a sequence of $l_{a_j}$ words $\{{x^{a_j}_1,x^{a_j}_2,...,x^{a_j}_{l_{a_j}}}\}$. And $l_{a_j}$ denotes the length of the ${j}$-th sentence ${a_j}$ in the answer $A_i$.

\subsection{Gated Recurrent Unit Encoder} 
For the sentence in the question and the answer, we follow previous work ~\cite{jang2019video, wang2020heterogeneous} and represent each word as a vector using the pre-trained GloVe word embedding denoted as $E = {\{e_i:i\leq l_{q}, e_i \in \mathcal{R}^{300}\}}$ and $F = {\{e_j:j\leq l_{a}, e_j \in R^{300}\}}$, where $l_q$ and $l_a$ indicate the number of embedded words in the sentence of question or answer. 

To obtain contextual representations to aggregate temporal information and enhance reasoning ability, we employ two independent Gated Recurrent Unit (GRU)~\cite{chung2014empirical} to each sentence in the questions and the answers separately. To reduce the cost of information in the longer answer, we set the encoder of the answer to bidirectional. The generated question sentence features $V_q\in\mathcal{R}^{l_q\times d}$ and answer sentence features $V_a\in\mathcal{R}^{l_a\times d}$ denote as:
\begin{equation}
    V_q, v_{l_q}=\overrightarrow{GRU}(E,\theta_{GRU}),
\end{equation}
\begin{equation}
    V_a,v_{l_a} = \overleftrightarrow {GRU}(F,\theta_{GRU}),
\end{equation}
\noindent where $v_{l_q}$ and $v_{l_a}$ are the outputs of the last hidden units which represent the global features of the two sentences in the question and the answer.

\subsection{Sentence-level GCN Encoder}
We propose the Sentence-level GCN Encoder module to capture the dependency information in the process of answering questions. Effectively modeling the context of the long-text answer requires capturing the inter-dependency and self-dependency among sentences in the question and the answer. Therefore, we construct a directed graph from the sequentially encoded sentences to capture the dependency between the question and the answer. Furthermore, we propose a neighborhood-based convolutional feature transformation process to create contextually features. The framework is detailed here.

First, we introduce the following notation: a single QA session is represented as a directed graph ${\mathcal{G}=(\mathcal{V},\mathcal{E},\mathcal{R},\mathcal{W})}$, with vertices $v_i\in \mathcal{V}$, labeld edges (relations) $r_{ij} \in \mathcal{E}$ where $r \in \mathcal{R}$ is the relation type of the edge between $v_i$ and $v_j$, and $\alpha_{ij}$ is the weight of the labeled edge $r_{ij}$, with $0 \leq \alpha_{ij} \leq 1$, where $\alpha_{ij} \in \mathcal{W}$ and $i$, $j$ $\in$ $[1,2,...,c_Q]$ and $[1,2,...,c_A]$, respectively. 

{\bf Graph Construction} The graph is constructed from the sentences in the following way.

{\bf Vertices:} Each sentences in the question or answer is represent as a vertex $v_i \in \mathcal{V}$ in $\mathcal{G}$. Each vertex $v_i$ is initialized with the corresponding sequentially encoded feature vector $h_i$, for all $i$ in $[1,2,...,c_Q]$ or $[1,2,...,c_A]$. We denote this representation vector as the vertex feature. Here, vertex features are subject to update downstream, when the transformation process is used to encode the context of QA pair.

{\bf Edges:} Construction of the edges $\mathcal{E}$ depends on the context of the current question and answer to be modeled. As mentioned, in this scenario, we need to evaluate several QA pairs in the interview session. And to acquire the ideal job opportunity, candidates would like to answer each sub-question more clearly, which leads to the answer text contains many sentences. It will be computationally quite expensive to construct the graph neural network through a full connection. Therefore, inspired by~\citet{ghosal2019dialoguegcn}, we employ a more efficient way to construct the edges by keeping a past context window size of ${p}$ and a future context window size of $f$. Hence, each vertex ${v_i}$ has an edge with the immediate $p$ sentences of the past: $v_{i-1},v_{i-2},...,v_{i-p}$, $f$ sentences of the future: $v_{i+1},v_{i+2},...,v_{i+f}$ and itself: $v_i$.

Meanwhile, as the graph is directed, vertices from the question or the answer or both can have edges in both directions with different relations.

{\bf Edge Weights:} We apply a similarity based attention module to acquire the edge weights. The attention function is computed in a way such that, for each vertex, the total weight of the incoming edges is 1. Considering the past and future context window size of $p$ and $f$, respectively, we calculated the weights as below,

\begin{equation}
\begin{split}
\alpha_{ij} &= softmax(g^T_iW_e[g_{i-p},...,g_{i+f}]),\\
& for\  j=i-p,...,i+f.
\end{split}
\label{eq-edgew}
\end{equation}
\noindent where $W_e$ represents the parameter to be learned. In this way, it can be ensured that the incoming edges of vertex $v_i$ receives a total weight contribution of 1.

{\bf Relations:} The relation $r$ of an edge $r_{ij}$ is designed in two aspects:

\emph{Internal temporal dependency} In the $i$-th question/answer, especially, the $i$-th answer, the relation depends on the relative position of occurrence of $u_i$ and $u_j$ in the question/answer: whether $u_i$ is appeared before $u_j$ or after. Therefore, there can be $c_{Q_i}$ (sentences in $Q_i$) * $c_{Q_i}$ + $c_{A_i}$ (sentences in $A_i$) * $c_{A_i}$ = ${c_{Q_i}}^2$ + ${c_{A_i}}^2$ relation types $r$ in the graph $\mathcal{G}$ for context understanding.

\emph{ Question-Answer interaction dependency} 
The relations also depends upon the interaction dependency between the question and answer in the QA session. To establish the latent semantic interaction relationships of question-to-answer and answer-to-question, we define the relation of Question-Answer interaction. There will be $c_{Q_i}$ * $c_{A_i}$ +$c_{A_i}$ * $c_{Q_i}$ = $2c_{Q_i}c_{A_i}$ relation types $r$ in the graph $\mathcal{G}$ for reasoning. Thus, the total number of distinct relation types in the graph $\mathcal{G}$ is ${c_{Q_i}}^2$ + ${c_{A_i}}^2$ + $2c_{Q_i}c_{A_i}$ = ${(c_{Q_i}+c_{A_i})}^2$.

\begin{table}[!ht]
\scriptsize
\centering
\begin{tabular}{|l|l|l|l|}
\hline
Relation  & $Q(u_i),A(u_j)$ & $i<j$ & $(i,j)$\\\hline 
1 & Q,Q & Yes & (1,2)\\
2 & Q,Q & No & (1,1),(2,1),(2,2) \\
3 & Q,A & - & (1,3),(1,4),(1,5),(2,3),(2,4),(2,5) \\
4 & A,Q & - & (3,1),(3,2),(4,1),(4,2),(5,1),(5,2)\\
5 & A,A & Yes & (3,4),(3,5),(4,5)\\
6 & A,A & No & (3,3),(4,3),(4,4),(5,3),(5,4),(5,5)\\\hline
\end{tabular}
\caption{\label{res-relationtype} $Q(u_i)$ and $A(u_j)$ indicate the source of sentence $u_i$ and $u_j$. The question and the answer in the current session imply ${(2+3)^2}=25$ distinct relation types. '-' means the connection between the question and the answer is independent of the index. The rightmost column denotes the indices of the vertices of the constituting edge which has the relation type indicated by the leftmost column.}
\end{table}

As we all know, the total assessment in an interview is affected by each QA session, and the evaluation of a single QA session should be based on the semantic interaction state between the question and the answer. Therefore, we hypothesize that explicit declaration of such relational edges in the graph would benefit in capturing the latent Relation of Question-Question (RQQ), Answer-Answer (RAA), and Question-Answer (RQA) among the QA session, which in succession would facilitate the total assessment of the whole interview.

As an illustration, let the question $Q$ and the answer $A$ in a QA session have 5 sentences, where $u_1,u_2$ are two sub-questions in $Q$, and $u_3,u_4,u_5$ are sentences in $A$. Then the edges and relations will be constructed as shown in Table~\ref{res-relationtype}.

\emph{Feature Transformation} 
The additive attention based sentence encoder mentioned the previous subsection provides effective sentence-level features $h^0_i$ for initialing the $i$-th node. Beyond that, the graph provides more dependency information between sentences in questions and answers. A more desirable way is to aggregate these information at the graph-level to get semantic status. A new feature vector $h^{(l+1)}_i$ is computed for vertex $v_i$ by aggregating local neighbourhood information through the relation specific transformation inspired from ~\cite{schlichtkrull2018modeling, ghosal2019dialoguegcn}:
\begin{equation}
\begin{split}
 h^{(l+1)}_i &=\sigma(\sum_{r\in \mathcal{R}} \sum_{j \in {N^r_i}}  \frac {\alpha_{ij}} {c_{i,r}} W^{(l)}_rg_j+\alpha_{ii} W^{(l)}_0g_i)\\
    & for\ i=1,2,...,N.
\end{split}
\end{equation}
\noindent where $\sigma$ is an activation function, and $W_0^{(1)}$ and $W_r^{1}$ are learnable parameters of the transformation. $\alpha_{ii}$ and $\alpha_{ij}$ are the weights of the edges, $N_i^r$ represents the neighbouring indices of vertex i under relation $r\in\mathcal{R}$. $c_{i,r}$ is the normalizer equal to $|N^r_i|$.

\subsection{Semantic-level GAT Encoder}
In order to reason on the semantic level, we employ a graph attention network to gather information from the nodes of the sentence-level graph and obtain the final hidden state $h'$ of each sentence. With the feature vector $h^{(l+1)}_i$ for initialing the $i$-th node in reasoning GAT, we can get the updated node representation $\tilde{h}^{(l+1)}_i$:
\begin{equation}
    \tilde{h}^{(l+1)}_i = \sum_{j\in N(i) }\tilde{\alpha}_{i,j} \tilde{W}^{(l)}h^{(l)}_i
\end{equation}
\noindent where $\tilde{\alpha}_{i,j}$ is the attention score between node $i$ and node $j$:
\begin{equation}
    \tilde{\alpha}_{i,j} = softmax(e_{ij}),
\end{equation}
\begin{equation}
    e_{ij} = \tilde{\sigma}(a^T[\tilde{W} \tilde{h}_i||\tilde{W}\tilde{h}_j])
\end{equation}
\noindent where $\tilde{\sigma}$ is the LeakyReLu activation function and $\tilde{W}$ denotes a learnable hyperparameter. After the sentence nodes are sufficiently updated on the semantic level, they are aggregated to a graph-level represention for the QA pair, Based on which the global representation of the interview can be obtained by a GRU encoder. We define the readout function as:
\begin{equation}
    h_{\mathcal G} = \frac {1} {|\tilde{\mathcal{V}}|} \sum_{i \in \tilde{\mathcal{V}}} \tilde{h}_i
\end{equation}

\subsection{Competency classifier}
Once the graph representation $h_{\mathcal G}$ of each QA pair is obtained, we feed it into a GRU encoder to capture the global representation $V_{final}$ of the interview. Then we feed it into a softmax layer to classify candidates:

\begin{equation}
    \hat y_{\mathcal{G}} = softmax(W_{final}V_{final}+b_{final}),
\end{equation}
\noindent where $W_{final}$ is a weight matrix and $b_{final}$ is the bias. As the problem we focused on the a binary classification, we apply the binary cross-entropy as our loss function.

\section{Experiments}
\subsection{Data and Metrics}
We evaluate our method on a real-world Chinese answer transcription (CHNAT) in the video job interview . The type of job is sales positions. And the answer transcriptions are obtained from an automatic speech recognition algorithm. Three experts are invited to annotate the same candidates, and we proceed with a majority vote to obtain the golden category. The CHNAT dataset split contains 2,313/289/290 candidates for training, validation, and test. To simplify the ASAT task, we set up a binary classification: based on the understanding of the textual answer, candidates who have been liked are considered part of the {\emph competent} class and others part of the incompetent class. Some statistics of the dataset are listed in Table~\ref{table-ds}. Although we are authorized by the candidates to use their interviews, the dataset will not be released to the public due to high privacy constraints.

\begin{table}[!ht]
\begin{center}
\begin{tabular}{|l|l|l|}\hline
Dataset & CHNAT\\ 
\hline
Training set & 2,313\\ 
Validation set & 289\\ 
Test set & 290\\ 
Questions per Candidate (mean) & 5.45\\ 
Hireable label propotion & 63.30\%\\ 
Total length & 507M\\ 
Length per question (mean) & 64\\ 
Length per answer (mean) & 256 \\
Vocab size & 21,128\\\hline 

\end{tabular}
\end{center}
\caption{\label{table-ds} Descriptive table of CHNAT: number of candidates in each set and overall statistics of the dataset}
\end{table}

Besides traditional evaluation metrics such as precision, recall, F1-score, and accuracy (ACC), we use the concordance correlation coefficient (CCC) proposed by ~\cite{lawrence1989concordance} to evaluate our model. In statistics, the concordance correlation coefficient measures the correlation and agreement between the predicted results of the model and the ground-truth distribution~\cite{deyo1991reproducibility, tzirakis2017end}.

\subsection{Settings and Hyper-parameters}
For fair comparisons with other methods, we take a consistent hyper-parameters to train our proposed model. We trained our baseline models for about 20 epochs - this is similar to the proposed HRGNN model. We limit the vocabulary to 21,128 and initialize tokens with 300-dimensional GloVe embeddings~\cite{pennington2014glove}. While training, we set the word embeddings to be trainable. We filter stop words and punctuations when creating sentence nodes and truncate the input sentence in the question and its corresponding answer to a maximum length of 50 and 295 separately. We set the batch size to 128, and initialize the GRU size with 50 and the attention size with 100. In RGCN and RGAT the sentence nodes with $d_s$=256 and edge features with $d_e=50$. We set the RGCN and RGAT layer to 1. And each RGAT layer is 16 heads. We applied dropout with a rate of 0.1. With the initial learning rate 0.001, learning rate decay 0.97, Adam optimizer~\cite{kingma2014adam} was used. Hyperparameters were optimized using grid search.

For all the experiments with HRGNN and our benchmark models, the scores (precision, recall, F1-score, ACC, and CCC) we present on the validation set and test set are mean values with 10 runs initialized by different random seeds.

\begin{table*}[!tp]
\centering
\begin{tabular}{|p{46mm}|p{7mm}p{7mm}p{7mm}|p{7mm}|p{7mm}|p{7mm}p{7mm}p{7mm}|p{7mm}|p{7mm}| }
\hline
 \multicolumn{1}{|c|}{\multirow{2}*{{Model}}}&\multicolumn{5}{c|}{\multirow{1}*{{ \textbf{Validation Set}}}} & \multicolumn{5}{c|}{\multirow{1}*{{ \textbf{Test Set}}}}\\

\cline{2-6} \cline{7-11} & P & R & F1 & ACC & CCC & P & R & F1 &ACC & CCC\\

\hline Non-seq~\cite{le2014distributed} & 76.00 & 83.06   &  79.37 & 72.66 & 39.10  & 76.60 & 78.69 & 77.63 & 71.38 &37.94 \\
 AGRU+FC  & 75.12 & 88.68 & 81.23 & 74.12 & 42.50 & 75.22 & 87.65  & 80.84 & 73.91 & 43.90\\

 HireNet~\cite{hemamou2019hirenet}  & 75.86 & 91.39 & 82.88 & 76.08 & 44.31 & 75.77 & 91.09  & 82.69 & 75.96 & 44.38\\

BERT+GRU  & \textbf{77.04} & 91.11 & 83.36 & \textbf76.99 & 46.73  & 76.31 & 90.99 & 82.87  & 76.38 & 45.43 \\

\hline HRGNN & 76.87 & \textbf{91.82}  &  \textbf{83.55} & \textbf{77.12} & \textbf{46.82}  & \textbf{78.25} & \textbf{91.47} & \textbf{84.25} & \textbf{78.49} & \textbf{50.78} \\

\hline
\end{tabular}
\caption{Automatic evaluation on CHNAT validation and test sets using precision (P), recall (R), F1, Accuracy (ACC) and CCC (Concordance Correlation Coefficien).}
\label{res-ZHNIAT}
\end{table*}

\subsection{Baselines}
In this section, we describe the baseline models in our experiments. 1) {\bf Non-sequential methods:} Similar to the state-of-the-art model--HireNet~\cite{hemamou2019hirenet} in AVIs, for the ASAT task, we first employ a non-sequential ({\bf Non-seq}, for abbreviation) model based on Doc2vec~\cite{le2014distributed} to represent the questions and answers, with three classic learning approaches (namely Ridge regression, Random Forest, and SVM) for classification. Best of the three approaches is shown. 2) {\bf Sequential methods:} Then two conventional neural network based models named {\bf AGRU+FC} and {\bf HireNet}~\cite{hemamou2019hirenet} are employed for comparison. {\bf AGRU+FC} is an intuitive baseline for ASAT. To obtain a better representation of the candidate's answer, An attention mechanism is utilized to extract the importance of each moment in the sequence representing the answer. Then, in the comparison with the known HireNet, to explore whether it is reasonable to encode question-answer pairs as a sequence, we employ a fully-connected classifier to make the final prediction. {\bf HireNet} is built relying on a hierarchical architecture. The low-level layer is constructed with an additive attention mechanism based GRU to encoder the local QA context, and the high-level layer consists of a global context encoder driven by another additive attention. With the hypothesis of the job titles are important for the job interview, HireNet includes vectors that encoder this contextual information. Due to we focus on the scoring of QA pairs in this work, we implement a variant of HireNet without the encoder for job titles. 3) {\bf BERT+GRU} As BERT~\cite{devlin2018bert} has achieved promising performance on several NLP tasks, we also implement one baseline method via fine-tuning BERT in the claim verification task. The GRU encoder in the low-level layer of HierNet is replaced by a BERT encoder. And in the high-level layer, we also employ GRU to encoder the question-answer pairs.


\subsection{Compared with Baseline Models}
We evaluated our proposed HRGNN on the validation and test set of CHNAT. To reduce the impact of randomness on HRGNN and baseline models, for all the experiments, the scores (precision, recall, F1-score, ACC, and CCC) we present are mean values with 10 runs initialized by different random seeds. Table~\ref{res-ZHNIAT} presents the performance of our model as well as the baselines. We observe that conventional neural network based sequential methods (AGRU+FC, HireNet) generally outperform Non-sequential models, suggesting that the sequential model benefits to the representation of questions and answers. Further, on the validation set and test test, HierNet with sequential GRU based classifier achieves (1.89\%, 2.14\%, and 1.91\% ) and (1.85\%, 2.05\%, and 0.48\%) of improvement in F1-score, ACC and CCC, respectively, compared to AGRU+FC model which employs fully-connected based classifier in the second stage of grading candidates. It indicates that it is reasonable to encode question-answer pairs as a sequence. Comparing the performance of HireNet and BERT+GRU, we can find that the pre-training based BERT+GRU model outperforms HireNet about (0.24\%, 0.73\%, and 2.32\% on the validation set) and (0.18\%, 0.42\%, and 1.03\% on the test set) in F1-score, ACC, and CCC, respectively. Intuitively, it shows that BERT has a stronger ability to represent text.

Besides, based on the results of all benchmark models, we can see that compared with the validation set, most of the performance of the non-sequential model, sequential model, and the fine-tuned BERT have variable degrees of decline in the test set. It may indicate that even if the model has a strong ability of text/context representation, it is difficult to transfer the ability to evaluate candidates from the validation set to the test set, without the deep interaction between questions and answers. On the test set, Compared to models which only driven by a sequential encoder, our model employs sentence-level GCN for constructing relational dependency and semantic-level GAT for QA reasoning, which brings (3.41\%, 4.58\%, and 6.88\%) and (1.56\%, 2.53\%, and 6.40\%) gains over AGRU+FC and HireNet in F1, ACC, and CCC, respectively. Although, the proposed HRGNN only outperforms pre-training based BERT+GRU about 0.11\%, 0.13\%, and 0.09\% in F1, ACC, and CCC separately. HRGNN achieves promising improvement against BERT+GRU on the test set. It indicates that the proposed model based relational dependency and QA reasoning has higher discrimination ability for unseen data. It further verifies the effectiveness of HRGNN.

\begin{figure}[!ht]
\centering
\centerline{ {\includegraphics[width=0.5\textwidth]{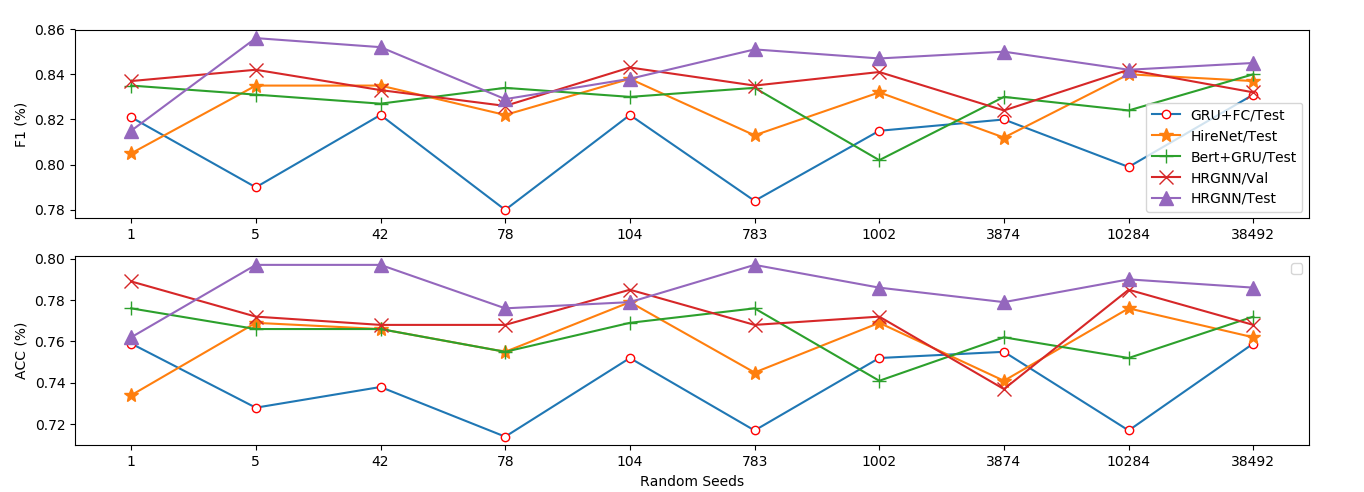}}}
\caption{F1-score and accuracy of different models with different random seeds on the test set. To analysis the transferability of HRGNN, we additionally plot a polyline of HRGNN on the validation set. "A/val" or "A/test" represents the performance of model A on the validation set or test set, respectively.}
\label{fig-random}
\end{figure}

Figure~\ref{fig-random} illustrates the F1-score and accuracy of different models with different random seeds. We can see that the most affected by random seed is AGRU+FC, followed by HireNet. On the test set, in most cases, HRGNN performs best and achieves the highest performance. Empirically, our proposed HRGNN was least affected by random seeds. As far as we know, it is almost impossible for the training set to contain all variable samples, models on the validation set tend to perform better than the test set. However, compared to the performance of HRGNN on the validation set (HRGNN/Val) and the test set (HRGNN/Test), we can observe that HRGNN/Test is higher than HRGNN/Val 7 times and 8 times in the F1-score and accuracy separately. It reveals that the semantic interaction and reasoning between questions and answers can improve the generalization ability of the evaluation model.

\subsection{Ablation Study}
We also perform ablation studies to assess the impact of different modules and different relation types on the model performance against text matching based model. Experimental results on the CHNAT dataset are shown in Table~\ref{res-as} and~\ref{res-rt}, respectively.

\begin{table}[!ht]
\scriptsize
\centering
\begin{tabular}{|l|lll|l|l|}
\hline
Model  & P & R & F1 & ACC & CCC \\\hline 

Baseline~\cite{hemamou2019hirenet}  & 75.86 & 91.39 & 82.88 & 76.08 & 44.31\\
+RGCN  & 76.17 & \textbf{92.01} & 83.22 & 76.51 & 45.10\\
+RGAT  & 77.30 & 90.40 & 83.30 & 77.02 & \textbf{47.20} \\

HRGNN  &  \textbf{76.87} & 91.82  &  \textbf{83.55} & \textbf{77.12} & 46.82 \\
\hline
Baseline~\cite{hemamou2019hirenet} &75.77 & 91.09  & 82.69 & 75.96 & 44.38\\
+RGCN  & 77.51 & 91.63 & 83.81 & 77.72 & 48.70\\
+RGAT  & 77.78 & 90.72 & 83.72 & 77.72 & 49.10\\
HRGNN  &\textbf{78.25} & 91.47 & \textbf{84.25} & \textbf{78.49} & \textbf{50.78} \\
\hline
\end{tabular}
\caption{\label{res-as} Ablation studies on the validation set and the test set of CHNAT using precision (P), recall (R), F1, Accuracy (ACC) and CCC (Concordance Correlation Coefficien). The upper and lower parts of the table correspond to the results of the validation set and the test set, respectively. We add each module separately and explore their influence on our model. '+' means we add the module to the baseline model.}
\end{table}

\begin{itemize}
    \item {\bf Impact of sentence-level RGCN.} When we add the sentence-level relational GCN to the baseline model (HireNet,~\cite{hemamou2019hirenet}), with the promotion of precision and recall, the F1-score of our model increases to 83.22\% (on the validation set) and 83.81\% (on the test set), which indicates the necessity of building relational graph to model the dependency between questions and answers. Particularly, on the test set, 
    the CCC increased from 44.38\% to 48.70\%, which shows that the proposed RGCN has a stronger capability to learn the logic of human scoring.
    \item {\bf Impact of semantic-level RGAT.} Using a reasoning graph attention network to encode the semantic-level information of the QA pair, performance in F1, ACC, and CCC increase to (83.30\%, 77.02\%, and 47.20\% on the validation set) and ( 83.72\%, 77.72\%, and 49.10\% on the test set), respectively, showing the contribution of semantic interaction over the QA pair.
    \item{\bf Impact of integration of two Graphs} When we employ the RGCN and RGAT on the top of the baseline method, HRGNN achieves the best performance in F1 and ACC. Meanwhile, we find that the improvement brought by the integration of RGCN and RGAT is not as high as that by a single module. We suspect that sometimes the relational dependency of the QA pair plays a similar role with the semantic interaction.
\end{itemize}

\begin{table}[!ht]
\scriptsize
\centering
\begin{tabular}{|l|lll|l|l|}
\hline

Realtion Type  & P & R & F1 & ACC & CCC \\\hline 
HRGNN &  76.87 & 91.82  &  \textbf{83.55} & \textbf{77.12} & \textbf{46.82} \\
-RQQ & \textbf{76.89} & 91.42 & 83.38 & 76.92 & 46.39 \\

-RAA  & 77.53 & 89.78 & 83.02 &76.75 & 46.59\\

-RQA & 74.89 & \textbf{92.84} & 82.80 & 75.57 & 42.16 \\

\hline
HRGNN &78.25 & 91.47 & \textbf{84.25} & \textbf{78.49} & \textbf{50.78} \\
-RQQ &77.91 & 91.37 & 84.00 & 78.07 & 49.75\\
-RAA & \textbf{78.34} & 90.27 &83.73 &77.93 & 49.80\\
-RQA & 76.46 &\textbf{92.35} &83.53 & 77.04&46.56 \\
\hline
\end{tabular}
\caption{\label{res-rt} Ablation studies of relation types on the validation set and the test set of CHNAT using precision (P), recall (R), F1, Accuracy (ACC), and CCC (Concordance Correlation Coefficient). We remove different relation types and explore their influence on our model. '-' means we remove the corresponding relation type from the original HRGNN.}
\end{table}

\begin{itemize}
    \item {\bf Impact of relation types.} To investigate the influence of relation types during the update process, we analyzed the performance of HRGNN under different relation types in Table~\ref{res-rt}. First, we can observe that when we remove the relation connection of sentences in the question, the results drop about (0.17\% and 0.2\% on the validation set) and (0.25\% and 0.42\% on the test set) in F1 and ACC, respectively. When we remove the relation type of RAA, the performance drops about (0.53\% and 0.37\%) and (0.52\% and 0.56\%) in F1 and ACC, which is invariably larger than the drop on -RQQ. The reason is that the answer contains more sentences and more semantic information. Further, when we remove the relation type of RQA, the performance drops more than that on -RAA. It suggests that the dependency between sentences in the question and the answer is more critical than that between the answer sentences for QA evaluation.
\end{itemize}

\section{Conclusion}
In this paper, we propose a hierarchical reasoning graph neural network (HRGNN) for the automatic scoring of answer transcriptions (ASAT) in the video job interview. The ASAT task is to score the competency of candidates based on several textual question-answer pairs. Unlike other matching based methods or frameworks, HRGNN can utilize the relational dependency of sentences in the questions and answers, and aggregate them in the semantic level with reasoning flow between different graph layers. Particularly, the proposed relational graph convolutional network (RGCN) module constructs internal temporal dependency and question-answer interaction dependency to represent the relations between sentences in the question and the answer. And in the graph-based reasoning part, we propose a graph attention network to further aggregate semantic interactions between the question and the answer. Finally, we apply a GRU-based classifier to discriminate the candidate is competent or not. Empirical results with 10 random seeds show that our model achieves state-of-the-art on a Chinese real-world dataset (CHNAT).

\bibliography{reference}

\begin{thebibliography}{42}
\providecommand{\natexlab}[1]{#1}
\providecommand{\url}[1]{\texttt{#1}}
\providecommand{\urlprefix}{URL }
\expandafter\ifx\csname urlstyle\endcsname\relax
  \providecommand{\doi}[1]{doi:\discretionary{}{}{}#1}\else
  \providecommand{\doi}{doi:\discretionary{}{}{}\begingroup
  \urlstyle{rm}\Url}\fi

\bibitem[{Battaglia et~al.(2018)Battaglia, Hamrick, Bapst, Sanchez-Gonzalez,
  Zambaldi, Malinowski, Tacchetti, Raposo, Santoro, Faulkner
  et~al.}]{battaglia2018relational}
Battaglia, P.~W.; Hamrick, J.~B.; Bapst, V.; Sanchez-Gonzalez, A.; Zambaldi,
  V.; Malinowski, M.; Tacchetti, A.; Raposo, D.; Santoro, A.; Faulkner, R.;
  et~al. 2018.
\newblock Relational inductive biases, deep learning, and graph networks.
\newblock \emph{arXiv preprint arXiv:1806.01261} .

\bibitem[{Bian et~al.(2019)Bian, Zhao, Song, Zhang, and Wen}]{bian2019domain}
Bian, S.; Zhao, W.~X.; Song, Y.; Zhang, T.; and Wen, J.-R. 2019.
\newblock Domain adaptation for person-job fit with transferable deep global
  match network.
\newblock In \emph{Proceedings of the 2019 Conference on Empirical Methods in
  Natural Language Processing and the 9th International Joint Conference on
  Natural Language Processing (EMNLP-IJCNLP)}, 4812--4822.

\bibitem[{Chen, Wu, and Zaki(2020)}]{chen2020toward}
Chen, Y.; Wu, L.; and Zaki, M.~J. 2020.
\newblock Toward Subgraph Guided Knowledge Graph Question Generation with Graph
  Neural Networks.
\newblock \emph{arXiv preprint arXiv:2004.06015} .

\bibitem[{Chung et~al.(2014)Chung, Gulcehre, Cho, and
  Bengio}]{chung2014empirical}
Chung, J.; Gulcehre, C.; Cho, K.; and Bengio, Y. 2014.
\newblock Empirical evaluation of gated recurrent neural networks on sequence
  modeling.
\newblock \emph{arXiv preprint arXiv:1412.3555} .

\bibitem[{CLAUDIA and CC-rater(2003)}]{claudia2003automated}
CLAUDIA, L.; and CC-rater, M. 2003.
\newblock Automated scoring of short-answer questions.
\newblock \emph{Computers and the Humanities} 37: 92--96.

\bibitem[{Conneau et~al.(2017)Conneau, Kiela, Schwenk, Barrault, and
  Bordes}]{conneau2017supervised}
Conneau, A.; Kiela, D.; Schwenk, H.; Barrault, L.; and Bordes, A. 2017.
\newblock Supervised learning of universal sentence representations from
  natural language inference data.
\newblock \emph{arXiv preprint arXiv:1705.02364} .

\bibitem[{Dai, Dai, and Song(2016)}]{dai2016discriminative}
Dai, H.; Dai, B.; and Song, L. 2016.
\newblock Discriminative embeddings of latent variable models for structured
  data.
\newblock In \emph{International conference on machine learning}, 2702--2711.

\bibitem[{Devlin et~al.(2018)Devlin, Chang, Lee, and
  Toutanova}]{devlin2018bert}
Devlin, J.; Chang, M.-W.; Lee, K.; and Toutanova, K. 2018.
\newblock Bert: Pre-training of deep bidirectional transformers for language
  understanding.
\newblock \emph{arXiv preprint arXiv:1810.04805} .

\bibitem[{Deyo, Diehr, and Patrick(1991)}]{deyo1991reproducibility}
Deyo, R.~A.; Diehr, P.; and Patrick, D.~L. 1991.
\newblock Reproducibility and responsiveness of health status measures
  statistics and strategies for evaluation.
\newblock \emph{Controlled clinical trials} 12(4): S142--S158.

\bibitem[{Ghosal et~al.(2019)Ghosal, Majumder, Poria, Chhaya, and
  Gelbukh}]{ghosal2019dialoguegcn}
Ghosal, D.; Majumder, N.; Poria, S.; Chhaya, N.; and Gelbukh, A. 2019.
\newblock Dialoguegcn: A graph convolutional neural network for emotion
  recognition in conversation.
\newblock \emph{arXiv preprint arXiv:1908.11540} .

\bibitem[{Goenka et~al.(2020)Goenka, Piplani, Sawhney, Mathur, and
  Shah}]{goenka2020esas}
Goenka, P.; Piplani, M.; Sawhney, R.; Mathur, P.; and Shah, R.~R. 2020.
\newblock ESAS: Towards Practical and Explainable Short Answer Scoring (Student
  Abstract).
\newblock In \emph{Proceedings of the AAAI Conference on Artificial
  Intelligence}, volume~34, 13797--13798.

\bibitem[{Hemamou et~al.(2019{\natexlab{a}})Hemamou, Felhi, Martin, and
  Clavel}]{hemamou2019slices}
Hemamou, L.; Felhi, G.; Martin, J.-C.; and Clavel, C. 2019{\natexlab{a}}.
\newblock Slices of Attention in Asynchronous Video Job Interviews.
\newblock In \emph{2019 8th International Conference on Affective Computing and
  Intelligent Interaction (ACII)}, 1--7. IEEE.

\bibitem[{Hemamou et~al.(2019{\natexlab{b}})Hemamou, Felhi, Vandenbussche,
  Martin, and Clavel}]{hemamou2019hirenet}
Hemamou, L.; Felhi, G.; Vandenbussche, V.; Martin, J.-C.; and Clavel, C.
  2019{\natexlab{b}}.
\newblock Hirenet: A hierarchical attention model for the automatic analysis of
  asynchronous video job interviews.
\newblock In \emph{Proceedings of the AAAI Conference on Artificial
  Intelligence}, volume~33, 573--581.

\bibitem[{Jang et~al.(2019)Jang, Song, Kim, Yu, Kim, and Kim}]{jang2019video}
Jang, Y.; Song, Y.; Kim, C.~D.; Yu, Y.; Kim, Y.; and Kim, G. 2019.
\newblock Video question answering with spatio-temporal reasoning.
\newblock \emph{International Journal of Computer Vision} 127(10): 1385--1412.

\bibitem[{Jiang et~al.(2019)Jiang, Zhang, Li, Bendersky, Golbandi, and
  Najork}]{jiang2019semantic}
Jiang, J.-Y.; Zhang, M.; Li, C.; Bendersky, M.; Golbandi, N.; and Najork, M.
  2019.
\newblock Semantic text matching for long-form documents.
\newblock In \emph{The World Wide Web Conference}, 795--806.

\bibitem[{Jiang and Han(2020)}]{jiang2020reasoning}
Jiang, P.; and Han, Y. 2020.
\newblock Reasoning with Heterogeneous Graph Alignment for Video Question
  Answering.
\newblock In \emph{AAAI}, 11109--11116.

\bibitem[{Kingma and Ba(2014)}]{kingma2014adam}
Kingma, D.~P.; and Ba, J. 2014.
\newblock Adam: A method for stochastic optimization.
\newblock \emph{arXiv preprint arXiv:1412.6980} .

\bibitem[{Lawrence and Lin(1989)}]{lawrence1989concordance}
Lawrence, I.; and Lin, K. 1989.
\newblock A concordance correlation coefficient to evaluate reproducibility.
\newblock \emph{Biometrics} 255--268.

\bibitem[{Le and Mikolov(2014)}]{le2014distributed}
Le, Q.; and Mikolov, T. 2014.
\newblock Distributed representations of sentences and documents.
\newblock In \emph{International conference on machine learning}, 1188--1196.

\bibitem[{Lun et~al.(2020)Lun, Zhu, Tang, and Yang}]{lun2020multiple}
Lun, J.; Zhu, J.; Tang, Y.; and Yang, M. 2020.
\newblock Multiple Data Augmentation Strategies for Improving Performance on
  Automatic Short Answer Scoring.
\newblock In \emph{AAAI}, 13389--13396.

\bibitem[{Luo et~al.(2019{\natexlab{a}})Luo, Zhang, Zhang, and
  Wu}]{luo2019improving}
Luo, W.; Zhang, C.; Zhang, X.; and Wu, H. 2019{\natexlab{a}}.
\newblock Improving Action Recognition with the Graph-Neural-Network-based
  Interaction Reasoning.
\newblock In \emph{2019 IEEE Visual Communications and Image Processing
  (VCIP)}, 1--4. IEEE.

\bibitem[{Luo et~al.(2019{\natexlab{b}})Luo, Zhang, Wen, and
  Zhang}]{luo2019resumegan}
Luo, Y.; Zhang, H.; Wen, Y.; and Zhang, X. 2019{\natexlab{b}}.
\newblock ResumeGAN: An Optimized Deep Representation Learning Framework for
  Talent-Job Fit via Adversarial Learning.
\newblock In \emph{Proceedings of the 28th ACM International Conference on
  Information and Knowledge Management}, 1101--1110.

\bibitem[{Marcheggiani, Bastings, and Titov(2018)}]{marcheggiani2018exploiting}
Marcheggiani, D.; Bastings, J.; and Titov, I. 2018.
\newblock Exploiting semantics in neural machine translation with graph
  convolutional networks.
\newblock \emph{arXiv preprint arXiv:1804.08313} .

\bibitem[{Mueller and Thyagarajan(2016)}]{mueller2016siamese}
Mueller, J.; and Thyagarajan, A. 2016.
\newblock Siamese recurrent architectures for learning sentence similarity.
\newblock In \emph{thirtieth AAAI conference on artificial intelligence}.

\bibitem[{Pan et~al.(2020)Pan, Xie, Feng, Chua, and Kan}]{pan2020semantic}
Pan, L.; Xie, Y.; Feng, Y.; Chua, T.-S.; and Kan, M.-Y. 2020.
\newblock Semantic Graphs for Generating Deep Questions.
\newblock \emph{arXiv preprint arXiv:2004.12704} .

\bibitem[{Pennington, Socher, and Manning(2014)}]{pennington2014glove}
Pennington, J.; Socher, R.; and Manning, C.~D. 2014.
\newblock Glove: Global vectors for word representation.
\newblock In \emph{Proceedings of the 2014 conference on empirical methods in
  natural language processing (EMNLP)}, 1532--1543.

\bibitem[{Qin et~al.(2018)Qin, Zhu, Xu, Zhu, Jiang, Chen, and
  Xiong}]{qin2018enhancing}
Qin, C.; Zhu, H.; Xu, T.; Zhu, C.; Jiang, L.; Chen, E.; and Xiong, H. 2018.
\newblock Enhancing person-job fit for talent recruitment: An ability-aware
  neural network approach.
\newblock In \emph{The 41st International ACM SIGIR Conference on Research \&
  Development in Information Retrieval}, 25--34.

\bibitem[{Saha et~al.(2019)Saha, Dhamecha, Marvaniya, Foltz, Sindhgatta, and
  Sengupta}]{saha2019joint}
Saha, S.; Dhamecha, T.~I.; Marvaniya, S.; Foltz, P.; Sindhgatta, R.; and
  Sengupta, B. 2019.
\newblock Joint Multi-Domain Learning for Automatic Short Answer Grading.
\newblock \emph{arXiv preprint arXiv:1902.09183} .

\bibitem[{Schlichtkrull et~al.(2018)Schlichtkrull, Kipf, Bloem, Van Den~Berg,
  Titov, and Welling}]{schlichtkrull2018modeling}
Schlichtkrull, M.; Kipf, T.~N.; Bloem, P.; Van Den~Berg, R.; Titov, I.; and
  Welling, M. 2018.
\newblock Modeling relational data with graph convolutional networks.
\newblock In \emph{European Semantic Web Conference}, 593--607. Springer.

\bibitem[{Shen et~al.(2018)Shen, Zhu, Zhu, Xu, Ma, and Xiong}]{shen2018joint}
Shen, D.; Zhu, H.; Zhu, C.; Xu, T.; Ma, C.; and Xiong, H. 2018.
\newblock A joint learning approach to intelligent job interview assessment.
\newblock In \emph{IJCAI}, 3542--3548.

\bibitem[{Suen, Hung, and Lin(2019)}]{suen2019tensorflow}
Suen, H.-Y.; Hung, K.-E.; and Lin, C.-L. 2019.
\newblock TensorFlow-based automatic personality recognition used in
  asynchronous video interviews.
\newblock \emph{IEEE Access} 7: 61018--61023.

\bibitem[{Sultan, Salazar, and Sumner(2016)}]{sultan2016fast}
Sultan, M.~A.; Salazar, C.; and Sumner, T. 2016.
\newblock Fast and easy short answer grading with high accuracy.
\newblock In \emph{Proceedings of the 2016 Conference of the North American
  Chapter of the Association for Computational Linguistics: Human Language
  Technologies}, 1070--1075.

\bibitem[{Tzirakis et~al.(2017)Tzirakis, Trigeorgis, Nicolaou, Schuller, and
  Zafeiriou}]{tzirakis2017end}
Tzirakis, P.; Trigeorgis, G.; Nicolaou, M.~A.; Schuller, B.~W.; and Zafeiriou,
  S. 2017.
\newblock End-to-end multimodal emotion recognition using deep neural networks.
\newblock \emph{IEEE Journal of Selected Topics in Signal Processing} 11(8):
  1301--1309.

\bibitem[{Veli{\v{c}}kovi{\'c} et~al.(2017)Veli{\v{c}}kovi{\'c}, Cucurull,
  Casanova, Romero, Lio, and Bengio}]{velivckovic2017graph}
Veli{\v{c}}kovi{\'c}, P.; Cucurull, G.; Casanova, A.; Romero, A.; Lio, P.; and
  Bengio, Y. 2017.
\newblock Graph attention networks.
\newblock \emph{arXiv preprint arXiv:1710.10903} .

\bibitem[{Wang et~al.(2020)Wang, Liu, Zheng, Qiu, and
  Huang}]{wang2020heterogeneous}
Wang, D.; Liu, P.; Zheng, Y.; Qiu, X.; and Huang, X. 2020.
\newblock Heterogeneous Graph Neural Networks for Extractive Document
  Summarization.
\newblock \emph{arXiv preprint arXiv:2004.12393} .

\bibitem[{Xu et~al.(2017)Xu, Zhu, Zhu, Li, and Xiong}]{xu2017measuring}
Xu, T.; Zhu, H.; Zhu, C.; Li, P.; and Xiong, H. 2017.
\newblock Measuring the popularity of job skills in recruitment market: A
  multi-criteria approach.
\newblock \emph{arXiv preprint arXiv:1712.03087} .

\bibitem[{Yao, Mao, and Luo(2019)}]{yao2019graph}
Yao, L.; Mao, C.; and Luo, Y. 2019.
\newblock Graph convolutional networks for text classification.
\newblock In \emph{Proceedings of the AAAI Conference on Artificial
  Intelligence}, volume~33, 7370--7377.

\bibitem[{Zhang et~al.(2020{\natexlab{a}})Zhang, Chen, Yang, Ramamurthy, Li,
  Qi, and Song}]{zhang2020efficient}
Zhang, Y.; Chen, X.; Yang, Y.; Ramamurthy, A.; Li, B.; Qi, Y.; and Song, L.
  2020{\natexlab{a}}.
\newblock Efficient probabilistic logic reasoning with graph neural networks.
\newblock \emph{arXiv preprint arXiv:2001.11850} .

\bibitem[{Zhang, Liu, and Song(2018)}]{zhang2018sentence}
Zhang, Y.; Liu, Q.; and Song, L. 2018.
\newblock Sentence-state lstm for text representation.
\newblock \emph{arXiv preprint arXiv:1805.02474} .

\bibitem[{Zhang et~al.(2020{\natexlab{b}})Zhang, Yu, Cui, Wu, Wen, and
  Wang}]{zhang2020every}
Zhang, Y.; Yu, X.; Cui, Z.; Wu, S.; Wen, Z.; and Wang, L. 2020{\natexlab{b}}.
\newblock Every Document Owns Its Structure: Inductive Text Classification via
  Graph Neural Networks.
\newblock \emph{arXiv preprint arXiv:2004.13826} .

\bibitem[{Zhao et~al.(2017)Zhao, Zhang, Xiong, Botelho, and
  Heffernan}]{zhao2017memory}
Zhao, S.; Zhang, Y.; Xiong, X.; Botelho, A.; and Heffernan, N. 2017.
\newblock A memory-augmented neural model for automated grading.
\newblock In \emph{Proceedings of the Fourth (2017) ACM Conference on Learning@
  Scale}, 189--192.

\bibitem[{Zhou et~al.(2019)Zhou, Han, Yang, Liu, Wang, Li, and
  Sun}]{zhou2019gear}
Zhou, J.; Han, X.; Yang, C.; Liu, Z.; Wang, L.; Li, C.; and Sun, M. 2019.
\newblock GEAR: Graph-based evidence aggregating and reasoning for fact
  verification.
\newblock \emph{arXiv preprint arXiv:1908.01843} .

\end{thebibliography}

\end{document}